\documentclass[runningheads]{llncs}
\usepackage{graphicx}
\usepackage{listings}
\usepackage{wrapfig}

\begin{document}
\title{CLAIMED, a visual and scalable component library for Trusted AI\thanks{Supported by IBM Center for Open Source Data and AI Technologies (CODAIT)}}
%
%
\author{Romeo Kienzler\inst{1} \and Ivan Nesic\inst{2}}
\authorrunning{Romeo Kienzler et al.}
%
\institute{IBM, Center for Open Source Data and AI Technologies (CODAIT)\\
\and{University Hospital of Basel, Department of Radiology and Nuclear Medicine\\
}}

\maketitle              
\begin{abstract}
Deep Learning models are getting more and more popular but constraints on explainability, adversarial robustness and fairness are often major concerns for production deployment. Although the open source ecosystem is abundant on addressing those concerns, fully integrated, end to end systems are lacking in open source.
Therefore we provide an entirely open source, reusable component framework, visual editor and execution engine for production grade machine learning on top of Kubernetes, a joint effort between IBM and the University Hospital Basel. It uses Kubeflow Pipelines, the AI Explainability360 toolkit, the AI Fairness360 toolkit and the Adversarial Robustness Toolkit on top of ElyraAI, Kubeflow, Kubernetes and JupyterLab. Using the Elyra pipeline editor, AI pipelines can be developed visually with a set of jupyter notebooks. We explain how we've created a COVID-19 deep learning classification pipeline based on CT scans. We use the toolkit to highlight parts of the images which have been crucial for the model’s decisions. We detect bias against age and gender and finally, show how to deploy the model to KFServing to share it across different hospital data centers of the Swiss Personalized Health Network.    

\end{abstract}
\section{Introduction}
Open source software for performing individual AI pipeline tasks are abundant, but the community lacks a fully integrated, trusted and scalable visual tool. Therefore we have built CLAIMED, the visual \textbf{C}omponent \textbf{L}ibrary for \textbf{AI}, \textbf{M}achine Learning, \textbf{E}TL and \textbf{D}ata Science which runs on top of ElyraAI capable of pushing AI pipelines of any kind to Kubernetes. Any containerized application can become a component of the library. CLAIMED has been released\footnote{https://github.com/elyra-ai/component-library} under the Apache v2 open source license. In the following we introduce the  open source components we are integrating in our current release, followed by an overview of different component categories paired with a description of exemplary components used in health care. This pipeline is also available in open source \footnote{https://github.com/cloud-annotations/elyra-classification-training/tree/developer\_article}.

\subsection{Containerization and Kubernetes}
Virtualization opened up a lot of potential for managing the infrastructure, mainly the ability to run different operating systems on the same hardware at the same time. Next step of isolation can be performed for each of the microservices running on the server, but instead of managing access rights and resources on the host operating system, we can containerize these in separate packages with their own environments. Practical effect of this is that we are running each of the microservices as if they have their own dedicated virtual machine, but without the overhead of such endeavour. This is accomplished by running containers on top of the host operating system. An example of the containerization platform is Docker.

Containerization made it possible to run a large number of containers, which introduced the need of their orchestration. This means something, and hopefully not someone, needs to constantly take care that the system is in the desired state, it needs to scale up or down, manage communication between containers, schedule them, manage authentications, balance the load etc. Although there are other options like Docker Swarm, Kubernetes is the market leader in this domain. It was donated to CNCF by Google, which means a lot of Google's know-how and years of experience went into it. The system can run on public, on-prem or on hybrid clouds. On-prem installation is very important for institutions dealing with sensitive data. For IBM, Kubernetes is also strategic, joining CNCF, having moved all Watson Services to Kubernetes and aquired RedHat, IBM is now 3rd largest comitter to Kubernetes.

\subsection{DeepLearning with TensorFlow}
It is the second incarnation of the Google Brain project's scalable distributed training and inference system named DistBelief \cite{tf}. It supports myriad of hardware platforms, from mobile phones to GPU/TPU clusters, for both training and inference. It can even run in browser on the client's side, without the data ever leaving the machine. Apart from being a valuable tool in research, it is also being used in demanding production environments. On a development side, representing machine learning algorithms in a tree-like structures makes it a very good expression interface. Lastly, on the performance vs usability side, both eager and graph modes are supported. Meaning debugging is much simpler in the first case, and if there is the need for speed, one can use the latter.

\subsection{Kubeflow}
Kubeflow is a machine learning pipeline management and execution system running as first class citizen on top of Kubernetes. Besides making use of Kubernetes scalability it allows for reproducible work as machine learning pipelines and the results and intermediate artifacts of their executions are stored in a meta data repository.

\subsection{ElyraAI}
ElyraAI started as a set of extensions for the JupyterLab ecosystem. Here we concentrate on the pipeline editor of ElyraAI which allows for expression of machine learning workflows using a drag'n'drop editor and send them for execution on top of different engines like Kubeflow or Airflow. This allows for non-programmers to read and understand but also create machine learning workflows. ElyraAI also supports visualizing such pipelines in the browser (e.g. from a github repository).

\subsection{JupyterLab}
JupyterLab is one of the most popular development environments for data science. Therefore we started to support JupyterLab first. But the pipeline editor of ElyraAI will be supported in other environments as well, VSCode being next on the list.

\subsection{AI Explainability}
Besides their stunning performance, deep learning models face a lot of resistance for production usage because they are considered to be a black box. Technically (and mathematically) deep learning models are a series of non-linear feature space transformations - sounds scary, but in other words, per definition it is very hard to understand the individual processing steps a deep learning network performs. But techniques exist to look over a deep learning model's shoulder. The one we are using here is called LIME\cite{lime}. LIME takes the existing classification model and permutes images taken from the validation set (therefore the real class label is known) as long as a misclassification is happening. That way LIME can be used to create heat maps as image overlays to indicate regions of images which are most relevant for the classifier to perform best. In other words, we identify regions of the image the classifier is looking at.

\begin{figure}
  \includegraphics[width=\linewidth]{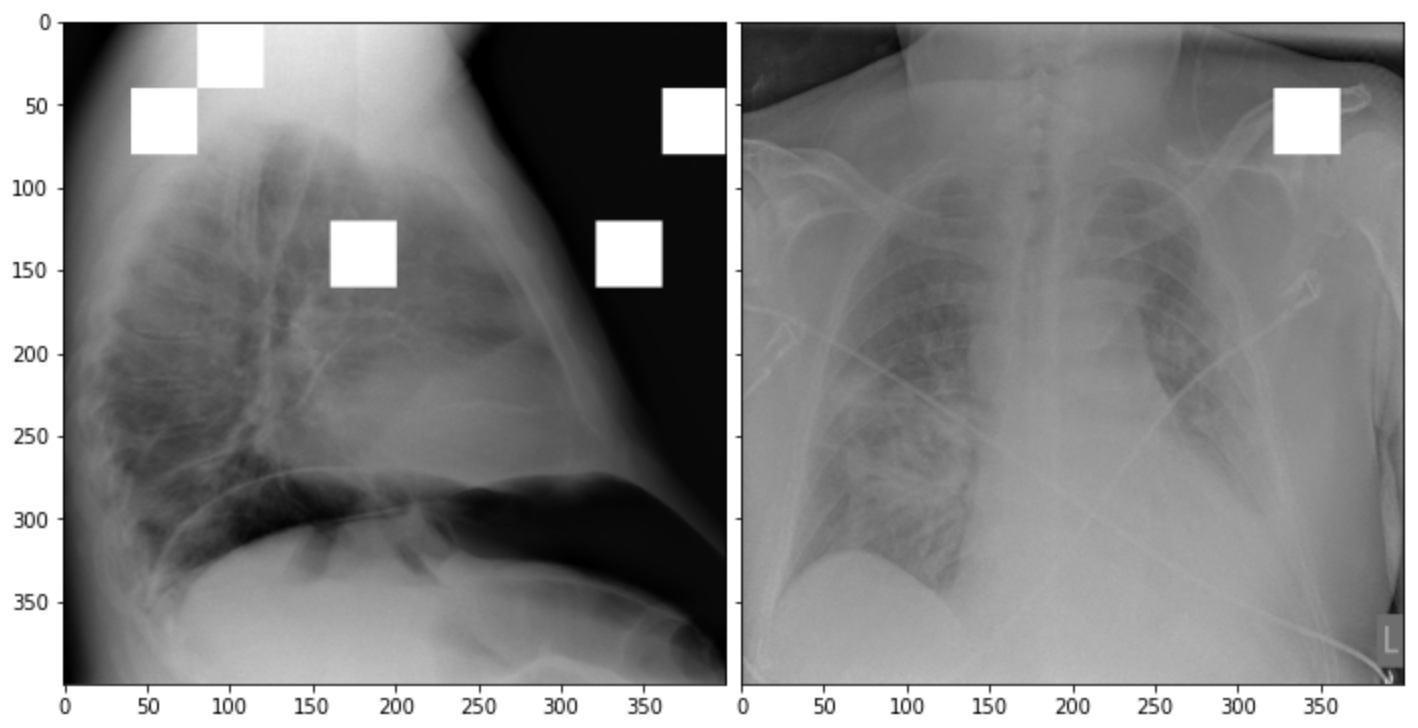}
  \caption{Example on how LIME helps to identify classification relevant areas of an image}
  \label{fig1}
\end{figure}

As Fig.~\ref{fig1}). illustrates, the most relevant areas in an image for classifying for COVID-19 are areas containing bones over lung tissue which indicates a problem with that particular classifier.

%

\subsection{AI Fairness and Bias}
So what is bias? Wikipedia says: "Bias is a disproportionate weight in favor of or against an idea or thing, usually in a way that is closed-minded, prejudicial, or unfair."\cite{wikipedia_bias} So here we have it. We want our model to be fair and unbiased towards protected attributes like gender, race, age, socioeconomic status, religion and so on. So wouldn't it be easy to just not "give" the model those data during training? It turns out that it isn't that simple. Protected attributes are often encoded in other attributes. For example, race, religion and socioeconomic status are latently encoded in attributes like zip code, contact method or types of products purchased. Fairness assessment and bias detection is an art on it's own. Luckily a huge number of single number metrics exist to assess bias in data and models. Here, we are using the AIF360\cite{aif360} library which IBM donated to the Linux Foundation AI and therefore is under open governance.

\subsection{AI Adversarial Robustness}
Another pillar of Trusted AI is adversarial robustness. As researchers found out, adversarial noise can be introduced in data (data poisoning) or models (model poisoning) to influence models decisions in favor of the adversarial. Libraries like the Adversarial Robustness Toolbox ART\cite{art} support all state-of-the-art attacks and defenses.

\section{System Implementation and Demo Use Case}
\subsection{A TrustedAI image classification pipeline}
\begin{figure}
  \includegraphics[width=\linewidth]{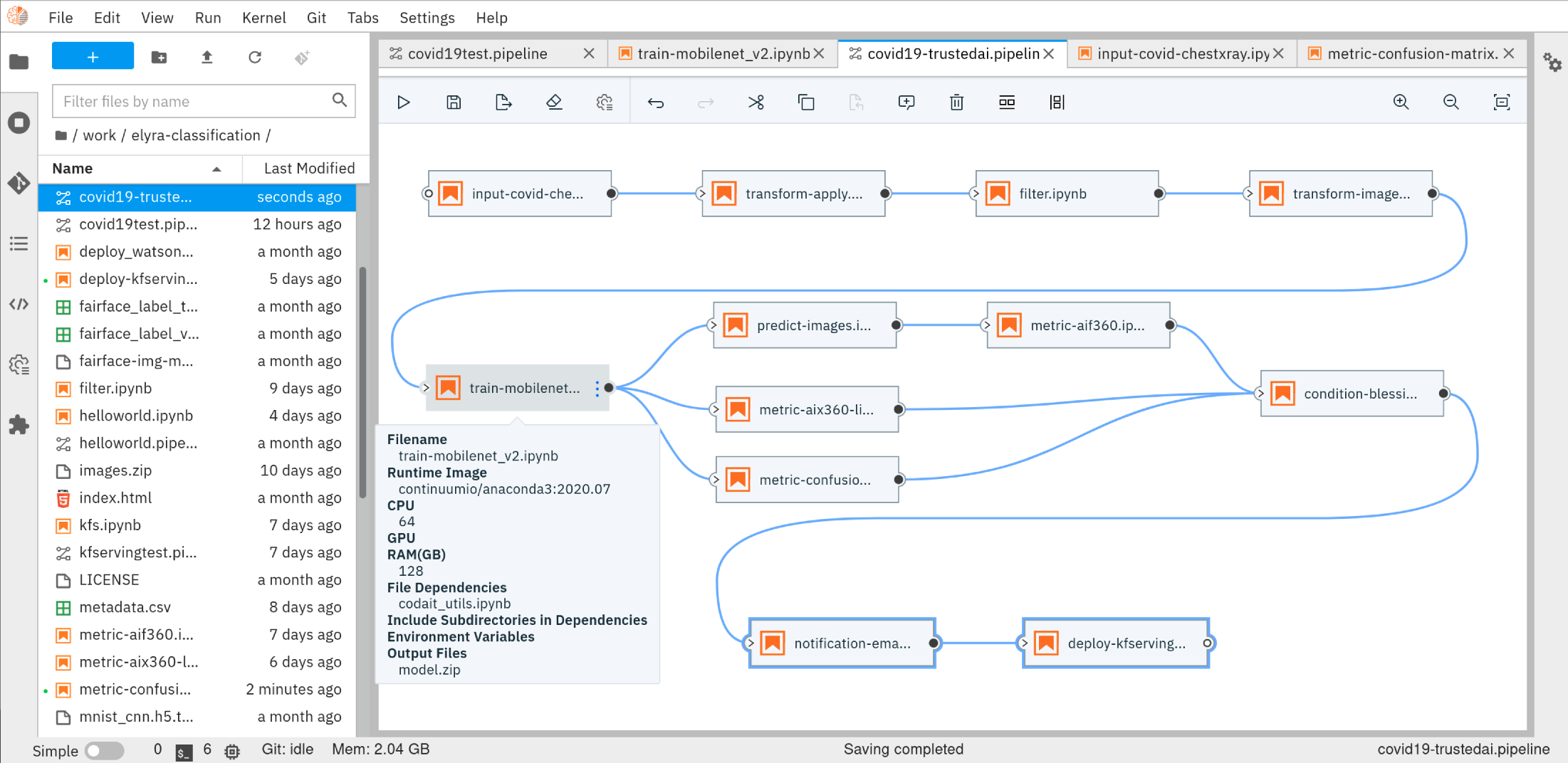}
  \caption{The exemplary TrustedAI pipeline for the health care use case}
  \label{pipeline}
\end{figure}

\begin{figure}
  \includegraphics[width=\linewidth]{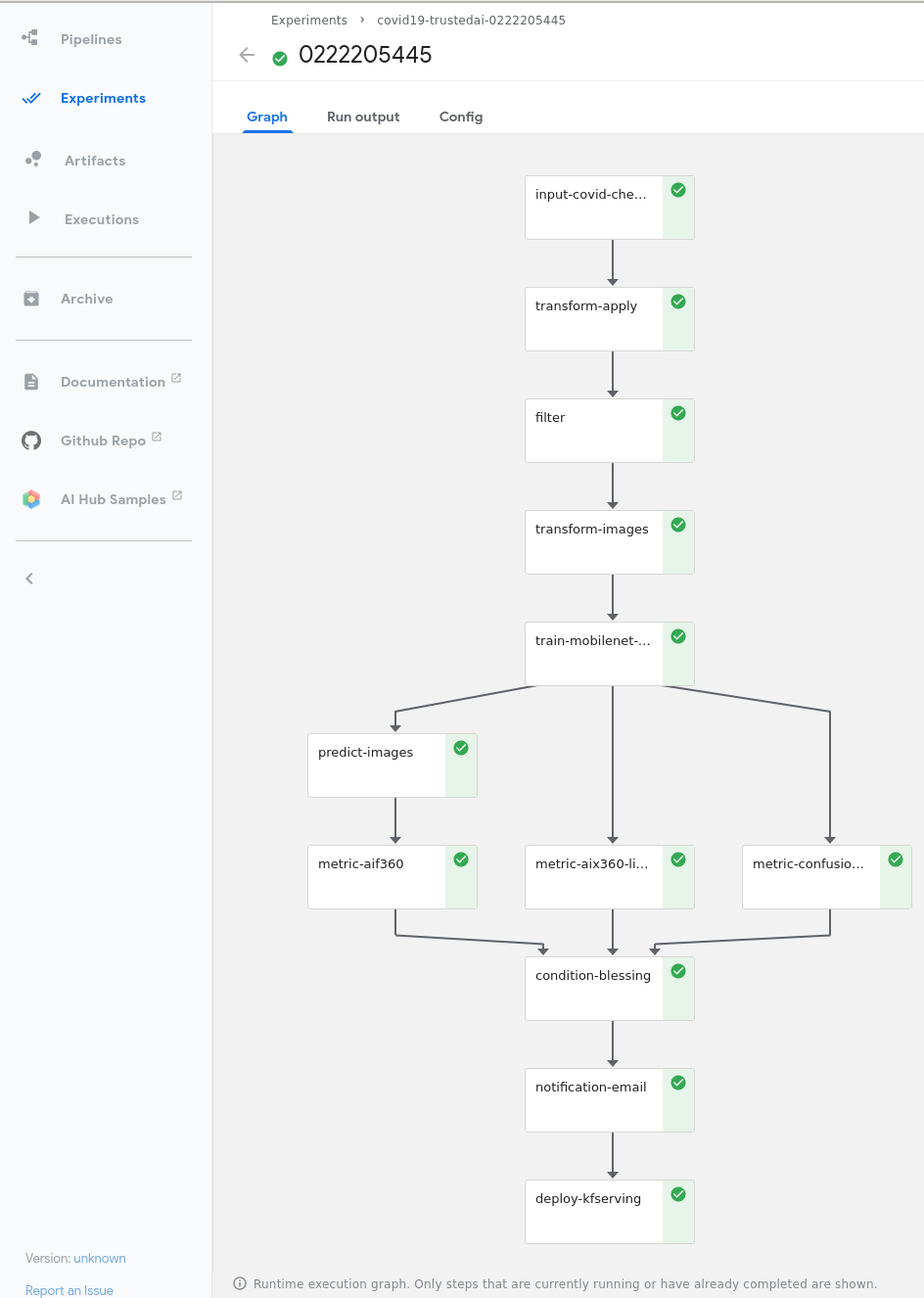}
  \caption{The pipeline once executed in Kubeflow}
  \label{kfp}
\end{figure}

As already mentioned previously, pipelines are a great way to introduce reproducibility, scaling, auditability and collaboration in machine learning. Pipelines are often a central part of a ML-Ops strategy. This especially holds for TrustedAI pipelines since reproducibility and auditability are even more important there. Figure \ref{pipeline} illustrates the exemplary TrustedAI pipeline we have built using the component library and figure \ref{kfp} is a screenshot taken from Kubeflow displaying the pipeline after finishing it's run.

\subsection{Pipeline Components}
In the following different categories of pipeline components are exemplified using components used in the Trusted AI image classification pipeline.

\subsubsection{Input Components}
In this particular case, we're pulling data directly from a GitHub repository via a public and permanent link. We just pull the metadata.csv and images folder. The component library will contain a component for each different type of data source like files and databases.

\subsubsection{Transform Components}
Sometimes, transformations on the metadata (or any other structured dataset) are necessary. Therefore, we provide a generic transformation component - in this case we just used it to change to format of the categories as the original file contained forward slashes which made it hard to use on the file system.
We just need to specify the column name and function to be applied on that column.

%

\subsubsection{Filter Components}
Similar to changing content of rows in a data set also removing rows is a common task in data engineering - therefore the filter stage allows for exactly that. It is enough to provide a predicate - in this case the predicate \verb|~metadata.filename.str.contains('.gz')| removes invalid images.

\subsubsection{Image Transformer Components}

\begin{wrapfigure}{l}{0.3\textwidth}
  \begin{center}
   \includegraphics[width=0.3\textwidth]{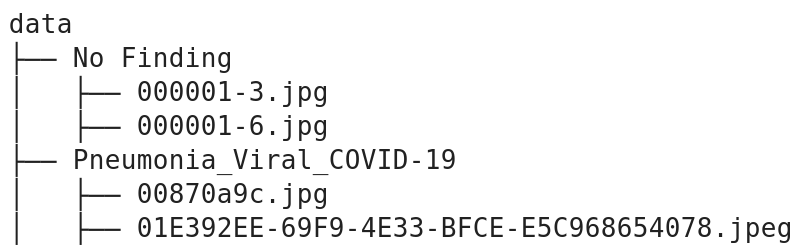}

  \end{center}
  \caption{de facto standard in folder structure for image classification data} \label{fig4}
\end{wrapfigure}

The de facto standard for labeled image data is putting images into one folder per class/category. But in this particular case, the raw data isn't in the required format. It's just a folder full of images and their properties are described in a separate CSV file. In addition to the class (or label) - finding in this case - this CSV file also contains information on the gender and age. So first, we just use the information on the finding label given in the CSV file and arrange the images in the appropriate folder structure, as illustrated in Fig.~\ref{fig4})

\subsubsection{Training Components}
\begin{wrapfigure}{l}{0.3\textwidth}
  \begin{center}
   \includegraphics[width=0.3\textwidth]{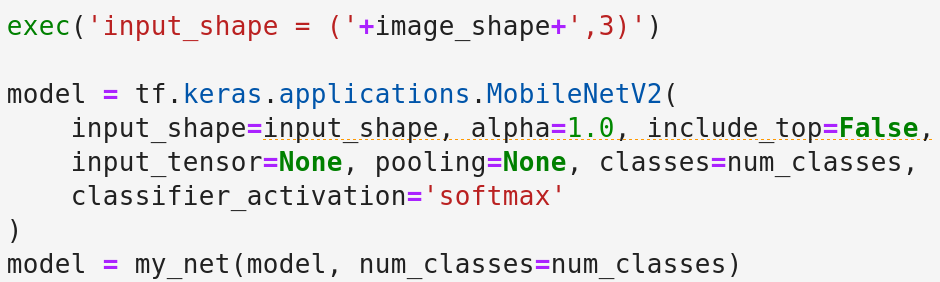}

  \end{center}
  \caption{Source code of the wrapped training component} \label{trainingstage}
\end{wrapfigure}

Understanding, defining and training deep learning models is an art on it's own. Training a deep learning image classification model requires a properly designed neural network architecture. Luckily, the community trends towards predefined model architectures, which are parameterized through hyper-parameters. At this stage, we are using the MobileNetV2, a small deep learning neural network architecture with the set of the most common parameters. It ships with the TensorFlow distribution - ready to use, without any further definition of neurons or layers. As shown in figure \ref{trainingstage}, only a couple of parameters need to be specified.

Although possible, hyper-parameter search is not considered in this processing stage as we want to make use of KubeFlow's hyper-parameter search capabilities leveraged through Katib\cite{katib} in the future.  

\subsubsection{Evaluation Components}
Besides define, compile and fit, a model needs to be evaluated before it goes into production. Evaluating classification performance against the target labels has been state-of-the-art since the beginning of machine learning, therefore we have added components like confusion matrix. But taking TrustedAI measures into account is a newly emerging practice. Therefore, components for AI Fairness, AI Explainability and AI Adversarial Robustness have been added to the component library.

\subsubsection{Blessing Components}
In Trusted AI it is important to obtain a blessing of assets like generated data, model or report to be published and used by other subsystems or humans. Therefore, a blessing component uses the results of the evaluation components to decide if the assets are ready for publishing.

\subsubsection{Publishing Components}
Depending on the asset type, publishing means either persisting a data set to a data store, deploying a machine learning model for consumption of other subsystems or publishing a report to be consumed by humans. Here, we exemplify this category by a KFServing component which publishes the trained TensorFlow deep learning model to Kubernetes. KFServing, on top of KNative, is particular interesting as it draws from Kubernetes capabilites like canary deployment and scalability (including scale to zero) in addition to built-in Trusted AI functionality.

\section{Future Work}
As of now, at least one representative component for each category has been released. Components are added to the library on a daily basis. The next components to be published are: Parallel Tensorflow Training with TFJob, Parallel Hyperparameter Tuning with Katib and Parallel Data Processing with Apache Spark. In addition, the next release of ElyraAI (v.2.3.0) will improve component's configuration options rendering capabilities, e.g. support for check-boxes and drop down menus and facilitated integration of exiting, containerized applications into the library without needing to wrap them in jupyter notebooks or python scripts.

\section{Conclusion} 
We've build and proposed a trustable, low-code, scalable and open source visual AI pipeline system on top of many de facto standard components used by the machine learning community. Using KubeFlow Pipelines provides reproducability and auditability. Using Kubernetes provides scalability and standardization. Using ElyraAI for visual development provides ease of use, such that all internal and external stakeholders are empowered to audit the system in all dimensions.

%
%
%

\begin{thebibliography}{8}
\bibitem{wikipedia_bias}
Wikipedia, \url{https://en.wikipedia.org/wiki/Bias}. Last accessed 18
Feb 2021

\bibitem{aif360}
AI Fairness 360 Toolkit, \url{https://github.com/Trusted-AI/AIF360}. Last accessed 18
Feb 2021

\bibitem{aix360}
AI Explainability 360 Toolkit, \url{https://github.com/Trusted-AI/AIX360}. Last accessed 18
Feb 2021

\bibitem{elyra}
Elyra AI, \url{https://github.com/elyra-ai}. Last accessed 18
Feb 2021

\bibitem{kubernetes}
Kubernetes, \url{https://kubernetes.io/}. Last accessed 18
Feb 2021

\bibitem{jupyterlab}
JupyterLab, \url{https://jupyter.org/}. Last accessed 18
Feb 2021

\bibitem{kfserving}
KFServing, \url{https://www.kubeflow.org/docs/components/serving/kfserving/}. Last accessed 18
Feb 2021


\bibitem{lime}
Marco Tulio Ribeiro and Sameer Singh and Carlos Guestrin: "Why Should {I} Trust You?": Explaining the Predictions of Any Classifier. Proceedings of the 22nd {ACM} {SIGKDD} International Conference on Knowledge Discovery and Data Mining, San Francisco, CA, USA, pp. 1135--1144 (2016)

\bibitem{katib}
Katib, \url{https://github.com/kubeflow/katib}. Last accessed 18
Feb 2021

\bibitem{tf}
TensorFlow: Large-scale machine learning on heterogeneous systems, white paper from \url{tensorflow.org}
Mart\'{\i}n~Abadi et al.


\bibitem{art}
Adversarial Robustness Toolbox, \url{https://github.com/Trusted-AI/adversarial-robustness-toolbox}. Last accessed 18
Feb 2021


\bibitem{ibmcncf}
IBM joining CNCF, \url{https://developer.ibm.com/technologies/containers/blogs/ibms-dedication-to-open-source-and-its-involvement-with-the-cncf/}. Last accessed 18
Feb 2021

\bibitem{cncf}
Cloud Native Computing Foundation, \url{https://www.cncf.io}. Last accessed 18
Feb 2021


\end{thebibliography}
%

\end{document}